\documentclass[letterpaper,twocolumn,10pt]{article}
\usepackage[dvipsnames]{xcolor}  
\usepackage{usenix}             

\usepackage{graphicx}            
\usepackage{amsmath,amssymb} 
\usepackage{algorithm}   
\usepackage[noend]{algpseudocode}       
\usepackage{booktabs}            
\usepackage{comment}             
\usepackage{pifont}              
\usepackage{multirow}            
\usepackage{makecell}            
\usepackage{subcaption}          
\usepackage{wrapfig}             
\usepackage[most]{tcolorbox}     
\tcbuselibrary{listings,skins}
\usepackage{placeins}            
\usepackage{tikz}  
\usepackage{bm}
\usepackage{url}
\usepackage{xurl}
\usepackage{hyperref}
\usepackage{dblfloatfix}

\usetikzlibrary{shapes.geometric,arrows,arrows.meta,positioning,fit,calc,backgrounds}
\algrenewcommand\algorithmiccomment[1]{\hfill\textit{/* #1 */}}

\newcommand{\PreCmt}[1]{\Statex \textit{/* #1 */}}
\newcommand{\PreCmtLong}[1]{%
  \Statex \parbox[t]{\dimexpr\linewidth\relax}{\textit{/* #1 */}}}

  \usepackage{paralist}
\newcommand{\xmark}{\ding{55}}%
\algnewcommand{\LineComment}[1]{\State \Comment{#1}}
\definecolor{EBBlue}{HTML}{1F78B4}

\definecolor{darkred}{RGB}{139,0,0}
\newcommand{\pbe}[1]{\vspace{0.75ex}\noindent{\bf\em #1}\hspace*{.3em}}
\newcommand{\ancomment}[1]{\textcolor{red}{\bf\small [ #1 --AN]}}

\newcommand{\eb}[1]{\textcolor{darkred}{\bf [E: #1]}}
\newcommand{\paragraphbe}[1]{\vspace{0.75ex}\noindent\textbf{\emph{#1}}\hspace*{.3em}}

\title{Throttling Web Agents Using Reasoning Gates}

\author{
  {\rm Abhinav Kumar}\quad
  {\rm Jaechul Roh}\quad
  {\rm Ali Naseh}\quad \vspace{0.1cm} \\
  {\rm Amir Houmansadr}  \quad
  {\rm Eugene Bagdasarian} \vspace{0.4cm} \\
  University of Massachusetts Amherst\\
  \texttt{\{abhinavk,jroh,anaseh,amir,eugene\}@cs.umass.edu}
}
\newcommand\blfootnote[1]{%
  \begingroup
  \renewcommand\thefootnote{}\footnote{#1}%
  \addtocounter{footnote}{-1}%
  \endgroup
}
\begin{document}
\maketitle 

\begin{abstract}
AI web agents use Internet resources at far greater speed, scale, and 
complexity---changing how users and services interact.
Deployed maliciously or erroneously, these agents could overload content
providers. At the same time, web agents can bypass CAPTCHAs and other defenses by mimicking user behavior or flood authentication systems with fake accounts. Yet providers must protect their services and content from denial-of-service attacks and scraping by web agents.

In this paper, we design a framework that imposes tunable costs on agents before providing access to resources; we call this \emph{Web Agent Throttling}.
We start by formalizing Throttling Gates as challenges issued to
an agent that are asymmetric, scalable, robust, and compatible with any agent.
Focusing on a common component---the language model---we require the agent to solve reasoning puzzles, thereby incurring excessive token-generation costs. However, we find that using existing puzzles, e.g., coding or math, as throttling gates fails to satisfy our properties.
To address this, we introduce \emph{rebus-based Reasoning Gates}, synthetic text puzzles that require multi–hop reasoning
over world knowledge (thereby throttling an agent's model). 
We design a scalable generation and verification protocol for such reasoning gates. 

Our framework achieves computational asymmetry, i.e., the response-generation cost is
9.2${\times}$ higher than the generation cost for SOTA models. 
We further deploy reasoning gates on a custom website and Model Context Protocol (MCP) servers and evaluate with real‐world web agents.
Finally, we discuss the limitations and environmental impact of real-world deployment of our framework.
\end{abstract}

\section{Introduction}


\begin{figure} [tbp]
    \centering
\includegraphics[width=\linewidth]{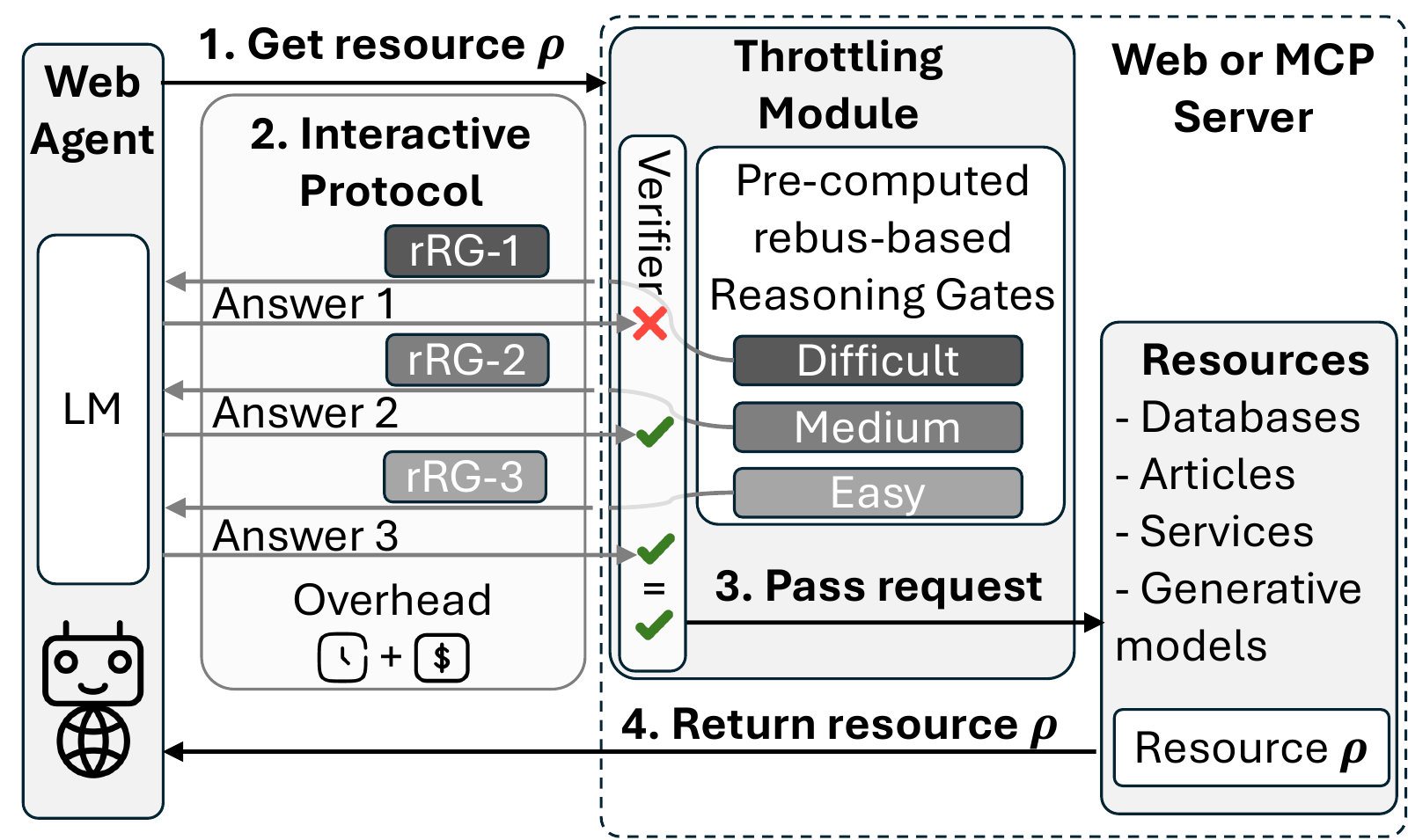}
    \caption{\textbf{Reasoning gates framework.}} 
    \label{fig:main}
\end{figure}

Internet resources are inherently limited, and many services struggle with denial-of-service (DoS) attacks~\cite{Neumann2000DenialofServiceA, mirkovic2004internet} and scraping~\cite{F5ScraperBots2025}.
Popularity of AI web agents~\cite{yehudai2025survey} that easily crawl the web while mimicking user behavior~\cite{flesch2025chatgptcrawler} exacerbate these threats. For example, Cloudflare recently accused Perplexity, an AI startup, of using anonymous web agents to intentionally bypass its anti-scraping defenses~\cite{cloudflare2025perplexity}. 
\blfootnote{Code available at \url{https://github.com/umass-aisec/throttling-web-agent}}
\blfootnote{Project Demo: \url{https://aisec.cs.umass.edu/demo/web-agent-throttling/}}

\makeatletter
\let\origthefootnote\thefootnote
\renewcommand\thefootnote{}
\let\thefootnote\origthefootnote
\makeatother



We argue that to prevent abuse and ensure equitable access, we should deploy mechanisms that enable web services to effectively \emph{throttle} emerging web agents, i.e., impose delays and costs on resource access.
Note that throttling the automated use of Internet resources is \emph{not} new;  Internet services have traditionally deployed CAPTCHAs~\cite{von2003captcha, motoyama2010re} to distinguish human users from automated bots. 
However, web agents powered by visual language models (VLMs) can easily bypass CAPTCHAs~\cite{deng2024oedipus, ding2025illusioncaptcha}. For instance, startups like Steel.dev~\cite{steel2024} have already commercialized CAPTCHA bypass solutions. 

Note that such throttling is \emph{complementary} to other agent controlling mechanisms. 
While user authentication and attestation mechanisms for users~\cite{whalen2022let, cloudflare2021cap} and new designs for agents~\cite{meunier2025webbotauth,simons2025futureoauth} will offer identity checks (if implemented), typical Internet services, e.g., public resources, cannot enforce selective access, while others will be prone to Sybil attacks~\cite{yuan2019detecting}. 
Therefore, completely blocking web agents is undesirable as many act on behalf of real users or services and provide legitimate utility. 
%

In this paper, we formalize the \emph{web agent throttling problem} as issuing a set of challenges, or \emph{throttling gates} that the agent has to solve before accessing the resource to discourage overuse or exploitation. We assert that throttling gates must satisfy four key criteria: (a) Computational Asymmetry: challenges must be expensive to solve but cheap to verify, (b) Scalability: task generation must be fast and not require manual curation, (c) Robustness: tasks must be resistant to web search, memorization, solver APIs, and gig workers, (d) Compatibility: human users, legitimate agents and low resource agents should not be unduly penalized. 
Although a single agent might not incur a significant overhead, when deployed at scale, as is common in DoS and scraping scenarios, these costs should deter companies and adversaries from abuse.

While AI web agents have different designs, the common part across them is a \emph{language model} (LM), therefore throttling challenges can only \emph{target them to impose costs}. Recent work shows that token generation for LMs is already costly and models are prone to overthinking~\cite{kumar2025overthinking, kaya2019shallow}. We establish that throttling gates should force the web agent to generate tokens, therefore exhausting its computational and/or financial resources. Therefore, we propose to use reasoning challenges, i.e. puzzles that cause LMs to generate tokens, for throttling calling them \emph{reasoning gates}.

We analyze a range of existing challenge formats (coding tasks~\cite{chen2021evaluating}, math puzzles~\cite{hendrycks2021measuring}, constraint problems~\cite{yang2023learning}, and trivia~\cite{rodriguez2019quizbowl}) and find that none meet all four criteria. Coding problems  lack scalable verification and are vulnerable to misuse; math and logic puzzles are difficult to auto-generate at varying difficulties; constraint problems can be outsourced to solver APIs; and trivia is easily solved via search. Even large-scale benchmarks like HLE~\cite{phan2025humanity} require coordination nearly 1,000 subject-matter experts to curate a test set. 

To overcome these limitations, we combine soft reasoning problems that use common knowledge~\cite{sprague2023musr} and rebus puzzles~\cite{gritsevskiy2024rebus} to create a scalable reasoning gates framework. We propose new multi-step, symbolic tasks that require LMs to combine commonsense, domain knowledge, multi-step reasoning, and symbolic composition capabilities. Each \textbf{rebus-based reasoning gate (rRG)} is derived from a random (word, domain) pair, allowing for low-cost verification and scalability without any LM generation based skew in the generated challenges. Importantly, solving them is difficult with basic web scraping or prompt engineering or human intervention, but manageable for capable LMs.

We introduce a lightweight throttling protocol that pre-generates rRGs offline and presents them to the agent when it requests the resources (see Figure~\ref{fig:main}).
Agents must solve these challenges to proceed, while the server bears minimal computational cost and can dynamically control the difficulty of the challenges. Our experiments reveal several key findings. First, we observe a compute-cost asymmetry of up to \textbf{9.2$\times$} between solving and generating the gates used in a session, making our approach efficient for providers while taxing for agents. Second, we show that smaller models (e.g., \texttt{o3-mini}) can generate gates that reliably challenge and slow down much larger models --- including state-of-the-art (SOTA) systems like o3 and Claude 3.7 without needing heavyweight infrastructure. 
Finally, in real-world tests using a live website with embedded gates, human participants struggled, taking 11 minutes to partially solve them. On the other hand,  Browser Use agent~\cite{browser_use2024} can solve all gates within 2-5 minutes while an agent can access the Model Context Protocol (MCP) server within 17 seconds.



\paragraph{Our Contributions.} In summary,  we (1) introduce and formalize the web agent throttling problem; (2) evaluate the limitations of conventional tasks as throttling gates; (3) propose and implement rRGs, a novel class of symbolic reasoning challenges designed for agent throttling; and (4) present a throttling protocol supported by extensive empirical studies, including real-world web agent evaluations.

\section{Background and Related Work}
\label{sec:background}

\pbe{AI Web Agents.} AI web agents can autonomously browse, interpret, and act on the web through multi-step, goal-driven interactions. Powered by multimodal models, they understand both text and images, while parsing page structure—including the DOM, layout, and semantic elements—to extract meaningful content~\cite{yang2025magma, yang2025agentic, sapkota2025ai}. These agents also integrate with external web tools such as browser APIs to execute tasks~\cite{zhou2023webarena, koh2024visualwebarena, deng2023mind2web, liang2024taskmatrix}. Recent advances in MCP~\cite{anthropicIntroducingModel} allow agents to connect with these tools securely by ensuring agents only access resources for which users have explicitly granted permission~\cite{cloudflare2025authorization}, and also made it possible to retrieve contextual data and orchestrate multi-tool tasks within real-world applications~\cite{hou2025model, hasan2025model}. Despite these advances, open challenges around robustness, trust, and orchestration remain, underscoring the need for secure and transparent infrastructure~\cite{ning2025survey}.



\pbe{Web-Scraping with Agents.} Companies increasingly deploy web agents to collect data for model training and user services~\cite{castle2024mapping, datadome2025llmcrawlers, nature2025botwars}. This has intensified debates around scraping, which threatens publisher revenues and complicates traffic detection~\cite{human2025scraping}. A notable case is the Perplexity–Cloudflare conflict, where crawler blocking led to evasive tactics like IP rotation and impersonation, countered by fingerprinting-based defenses~\cite{cloudflare2025perplexity, perplexity2025response}. Meanwhile, adaptive scrapers~\cite{huang2024autoscraper} and defenses leveraging watermarking and fingerprinting~\cite{liu2024expshield, mudryi2025hidden} illustrate an escalating arms race between sophisticated agents and detection systems.

\pbe{Denial-of-Service Attacks (DoS).} DoS attacks or Distributed-Denial-of-Service (DDoS) attacks overwhelm a system with excessive or malicious requests, rendering it inaccessible to legitimate users~\cite{gu2007denial, crosby2003denial, bellardo2003802}. More recently, this concept has expanded to target the efficiency and robustness of models such as LMs and VLMs~\cite{gao2024denial}. Adversarial inputs have been shown to induce disproportionate energy consumption and latency, significantly slowing down both cloud-based and on-device services. In particular, verbose or carefully crafted data can systematically disrupt VLMs, while large-scale studies confirm that LMs remain broadly vulnerable to resource exhaustion and latency-based poisoning~\cite{gao2024inducing, gao2024denial}. 


\pbe{Authentication and Attestation.} Agents may be registered under a company or directly to a user account, enabling service providers to authenticate them before granting access~\cite{south2025authenticated}. Various mechanisms enable such authentication~\cite{sakimura2014openid, hardt2012oauth, shingala2019json}, with stronger guarantees offered by attestation schemes~\cite{FIDOAttestation2024}, where devices prove authenticity through cryptographically signed, hardware-rooted assertions. On top of that, rate-limiting can block agents that send too many requests~\cite{wong2005effectiveness} or add time locks~\cite{abadi2023scalable}. 
Yet, these mechanisms remain vulnerable to large-scale threats, as fake accounts, botnets, or Sybil attacks can still overwhelm systems despite valid registration~\cite{yuan2019detecting, cao2012aiding}. Our throttling approach complements rather than replaces these efforts, and can be cascaded alongside them, as discussed in Section~\ref{sec:problem}.

\pbe{CAPTCHAs.} A longstanding defense against automated traffic is the CAPTCHA~\cite{von2003captcha, searles2023empirical}, which exploits perceptual and cognitive differences to distinguish humans from machines. Yet specialized deep learning models achieve over 83\% accuracy on reCAPTCHA v2~\cite{hossen2020object} and up to 94.25\% on tasks like street signs~\cite{zhou2018breaking}, while general-purpose VLMs now solve unseen CAPTCHA types without retraining~\cite{teohcaptchas}. Audio CAPTCHAs are likewise broken with high accuracy~\cite{bock2017uncaptcha}, and human-operated CAPTCHA farms further enable bypass~\cite{nakatsuka2021cacti}. As multimodal, reasoning-capable LMs integrate CoT reasoning, vision, and external solvers, the effectiveness of novel puzzle designs has been fundamentally undermined~\cite{gao2021captcha, deng2024oedipus, li2025attacks}. Finally, CAPTCHAs, including recent diffusion-based ones~\cite{ding2025illusioncaptcha}, are using multi-modal reasoning and are not compatible with text-only API or MCP servers.

\pbe{Puzzle-based defenses and Proof-of-Work.} Proof-of-Work (PoW) mechanisms throttle automated access by requiring moderately hard but feasible computation before service use~\cite{dwork2005pebbling}. Long used against DoS and DDoS attacks, puzzle-based defenses impose computational costs that deter large-scale abuse~\cite{wang2008multi, he2024cryptographic}. Multi-layered frameworks and cryptographic puzzles ensure that users, services, or agents must solve challenges before accessing resources, with adjustable difficulty enabling dynamic load regulation~\cite{neyyan2012game, Keller2018Consensus}. We analyze these approaches in Section~\ref{sec:existing} and argue they remain insufficient for throttling modern agents.

\section{Web Agent Throttling Problem}
\label{sec:problem}


As discussed in Section~\ref{sec:background}, authentication and attestation alone are not sufficient to prevent DoS attacks or large-scale unauthorized web scraping. Therefore, we advocate for an additional mechanism that can operate alongside these existing techniques, cascading current access-rate control to ensure resilience against malicious or excessive agent activity. We call this mechanism \emph{web agent throttling}, whose primary goal is to provide content providers with mechanisms that allow them to control the rate of automated access requests, thereby imposing cost or delay on high-volume activity while preserving access for legitimate users and benign automation. While some simple techniques such as fixed time delays can be considered throttling approaches, these methods are not suitable in practice, as discussed in Section~\ref{sec:background}. Instead, we introduce \emph{throttling gates}—points at which agents must solve one or more puzzles before being granted access. In the following subsections, we first define throttling gates and then identify the key properties that an effective mechanism must satisfy.





\subsection{Problem Definition}
We consider an agent $\mathcal{A}$ requesting resource $\rho$ from a service provider $\mathcal{S}$. To regulate access, $\mathcal{S}$ places $\mathcal{A}$ at a \emph{throttling gate}. A gate presents one or more puzzles that $\mathcal{A}$ must solve before proceeding. A gate may consist of a single puzzle or a sequence of puzzles (i.e., an interactive session). For each puzzle $c$ with solution $z$, the agent produces a response $r$. Access is granted if the response satisfies the correctness criteria defined by the provider—this may require exact equality ($r = z$) or partial correctness according to a specified threshold. These puzzles are generated by a puzzle generation algorithm $\mathcal{G}$, which produces a set of challenges $C_b$ that can be served to agents. The design goal is to ensure that gates remain inexpensive for $\mathcal{S}$ to generate and verify, while requiring significantly higher effort for $\mathcal{A}$ to solve.



\subsection{Threat Model} 
We assume an adversary who controls one or more automated AI agents equipped with at least an LM, giving them the minimum capability needed to attempt solving throttling puzzles. The adversary knows the overall design of the throttling gate, but does not have access to random seeds of the throttling mechanism and cannot directly modify or manipulate the gate construction process. Their primary goal is either to orchestrate a DoS attack against the target service provider by overwhelming it with large volumes of requests, or to conduct large-scale unauthorized scraping of the provider’s content. 

\subsection{Throttling Criteria} \label{sec:properties}



We now outline the essential properties that throttling challenges must satisfy to impose meaningful cost on automated agents while remaining efficient and accessible.


\pbe{Computational Asymmetry (A).}
An effective throttling gate should create a strong imbalance between the ease of challenge generation and verification versus the difficulty of solving it. This asymmetry ensures that content providers incur only negligible costs to issue and verify challenges, while automated agents are forced to expend substantially more resources to solve them, making large-scale abuse economically and computationally inefficient:
\begin{itemize}
    \item \textbf{A1: Generation should be cheap.} The process of generating reasoning problems should be fast and efficient, imposing minimal computational cost on the service provider and should allow for large scale generation.
    \item \textbf{A2: Verification should be constant-time.} The correctness of submitted answers should be verifiable instantly, adding no noticeable verifier end latency during the interactions.
    \item \textbf{A3: Solution should be expensive.} Solving the problem should require effort and cost from the agent, making it more expensive to send large-scale requests.
\end{itemize}


\pbe{Scalability (S).}
Since malicious web agents operate at massive scale, an effective throttling mechanism must be able to handle billions of requests without creating bottlenecks or imposing prohibitive costs on providers. To this end, scalability requires:
\begin{itemize}
    \item \textbf{S1: Billions of different problems.} The reasoning challenges should be easy to generate and highly diverse, enabling the service provider to issue unique problems per request and reduce the chance of memorization or reuse.
    \item \textbf{S2: Low operational costs.} The entire challenge pipeline—including generation, storage, selection, and serving—must remain lightweight and fast, ensuring that problem handling at inference time does not introduce overhead or bottlenecks.
\end{itemize}

\pbe{Robustness to Adversaries (R).}
A throttling mechanism is only effective if adversaries are forced to pay its intended computational cost. In practice, attackers will try either to bypass challenges cheaply or to exploit the defense itself. Thus, robustness requires that challenges cannot be trivially circumvented and do not create new vulnerabilities for the service provider:
\begin{itemize}
    \item \textbf{R1: Resistance to cheap shortcuts.} Challenges should not be easily solvable using off-the-shelf models, simple web searches, or other inexpensive tools. They must also be designed so that outsourcing to low-paid human workers is impractical, due to the difficulty, diversity, and range of potential problems.
    \item \textbf{R2: Server-side attacks.} While challenges should impose significant cost on agents, they must not require complex server-side execution that opens additional vulnerabilities. For example, code execution or containerized verification can expose software exploits, while relying on LM-based judges risks prompt-injection attacks or overthinking attacks~\cite{kumar2025overthinking}, where adversaries deliberately trigger excessive computation on the server.
\end{itemize}

\pbe{Compatibility (C).}
Throttling challenges must consider the diversity of deployment scenarios and agent capabilities. Without compatibility, throttling mechanisms risk either excluding benign agents or unfairly penalizing low-resource agents built on smaller LMs. Therefore:
\begin{itemize}
    \item \textbf{C1: Adjustable complexity.} The throttling gate should allow content providers to tune challenge difficulty, ensuring that benign agents can reliably pass while only high-volume adversarial activity incurs significant cost. This adaptability prevents weaker but legitimate agents from being excluded.
    \item \textbf{C2: Support for text-only, low-resource agents.} Challenges should be designed for agents that operate solely on textual inputs, since many low-capability agents do not support additional modalities or tools. Restricting puzzles to language-based formats avoids excluding such agents, reduces deployment complexity, and ensures that compatibility does not depend on extra software or multimodal capabilities.
\end{itemize}

The properties outlined above capture the necessary conditions for an effective throttling gate. They are not, however, sufficient on their own: additional design choices may be required to ensure security and practicality in deployment. Moreover, these properties are not binary. A given approach may exhibit partial robustness, scalability, or compatibility, but still fall short of achieving the level of asymmetry needed to meaningfully throttle automated agents. In the next section, we examine existing throttling strategies and show that they fail to satisfy even these necessary conditions, motivating the need for a new design that explicitly incorporates them.

\begin{table*}[tbp]
  \centering
  \begingroup
    \small
    \renewcommand{\arraystretch}{0.9}  
    \setlength{\extrarowheight}{2pt}
    \setlength{\tabcolsep}{3pt}

    \caption{\textbf{Comparison of throttling gates}. We measure existing and proposed rebus-based reasoning gates across four key properties essential for mitigating high-volume automated access: \textit{Computational Asymmetry} (A1: cheap generation, A2: constant-time verification, A3: expensive solution), \textit{Scalability} (S1: billions of different problems, S2: low operational overhead), \textit{Robustness} (R1: resistance to shortcuts, R2: no new server-side attacks), and \textit{Compatibility} (C1: adjustable complexity, C2: accessible to text-only and low-resource agents).}
    \label{tab:throttling-comparison}

    \begin{tabular*}{\textwidth}{@{\extracolsep{\fill}}@{}l l ccc cc cc cc@{}}
      \toprule
      \multirow{2}{*}{\textbf{Category}} &
      \multirow{2}{*}{\textbf{Challenge Type}} &
        \multicolumn{3}{c}{\makecell{\textbf{Computational}\\[-3pt]\textbf{Asymmetry}}} &
        \multicolumn{2}{c}{\textbf{Scalability}} &
        \multicolumn{2}{c}{\textbf{Robustness}} &
        \multicolumn{2}{c}{\textbf{Compatibility}} \\
      \cmidrule(lr){3-5} \cmidrule(lr){6-7} \cmidrule(lr){8-9} \cmidrule(lr){10-11}
      & & \textbf{A1} & \textbf{A2} & \textbf{A3}
        & \textbf{S1} & \textbf{S2}
        & \textbf{R1} & \textbf{R2}
        & \textbf{C1} & \textbf{C2} \\
      \midrule
      
      \multirow{2}{*}{\textbf{Classic Gates}} 
        & CAPTCHAs~\cite{von2003captcha}
        & \checkmark & \checkmark & \checkmark
        & \checkmark & \checkmark
        & \xmark     & \checkmark
        & \checkmark & \xmark \\

        & Proof-of-Work~\cite{dwork2005pebbling}
        & \checkmark & \checkmark & \checkmark
        & \checkmark & \checkmark
        & \xmark     & \checkmark
        & \checkmark     & \xmark \\


      \midrule
      
      \multirow{4}{*}{\textbf{Reasoning Gates}}
        & Solver-based~\cite{pan2023logic,feng2023language}
        & \checkmark   & \checkmark & \checkmark
        & \checkmark & \checkmark
        & \xmark     & \checkmark
        & \checkmark & \checkmark \\

        & Coding-based~\cite{chen2021evaluating}
        & \xmark   & \xmark     & \xmark
        & \checkmark & \xmark
        & \xmark     & \xmark
        & \checkmark & \checkmark \\

        & Advanced Sciences~\cite{hendrycks2021measuring}
        & \xmark   & \xmark     & \checkmark
        & \xmark     & \xmark
        & \checkmark & \xmark
        & \checkmark & \checkmark \\

        \cmidrule{2-11}
        & \textbf{\textit{rRG} (Ours)}
        & \textbf{\checkmark} & \textbf{\checkmark} & \textbf{\checkmark}
        & \textbf{\checkmark} & \textbf{\checkmark}
        & \textbf{\checkmark} & \textbf{\checkmark}
        & \textbf{\checkmark} & \textbf{\checkmark} \\
      \bottomrule
    \end{tabular*}
    \label{tab:property_comparison}
  \endgroup
\end{table*}



\section{Throttling Strategies}
\label{sec:existing}

Before introducing our proposed reasoning-gate mechanism, it is important to understand why both naive countermeasures and existing challenge-based approaches fail to effectively throttle modern LM-powered web agents. We evaluate them against the four key properties outlined in Section~\ref{sec:properties} and illustrated in Table~\ref{tab:property_comparison}.

\subsection{Classic Gates}\label{sec:existing_approaches}





\pbe{CAPTCHAs.} Although, CAPTCHAs can be broken as discussed in Section~\ref{sec:background}, could be used as throttling gates to slow down the agent.
From a computational asymmetry (A) perspective, they are cheap to generate (A1) and easy to verify (A2), while also taking time to solve (A3). 
Furthermore, CAPTCHAs can be easily generated in large numbers (S1) with relatively low operational cost (S2)~\cite{teohcaptchas}. However, their diversity is limited to a fixed set of puzzle types, which are vulnerable to memorization (R1), including training models on existing CAPTCHA, using human farms and other tools~\cite{wu2025mca, ding2025illusioncaptcha,zhou2018breaking,noury2020deep}. CAPTCHAs do not create new attack surfaces for the server (R2) as verification is easy. Finally, while CAPTCHAs allow some tuning of difficulty (C1), their dependence on non-textual modalities undermines accessibility for text-only agents, e.g. using API or MCP tools (C2). Overall, while easy to deploy, CAPTCHAs are not robust to bypasses and not compatible with text-only agents.

\pbe{Proof-of-Work (PoW).} PoW puzzles were designed to impose cost on clients~\cite{ali2020foundations, bar2025proof}, but they misalign with the needs of throttling web agents. These challenges are cheap to generate (A1) and trivial to verify (A2), and expensive to solve (A3)~\cite{biryukov2017equihash}. With respect to scalability (S), PoW challenges, like hash inversion, can be easily generated and distributed (S1) and efficiently deployed (S2). 
However, they do not guarantee uniformly expensive solutions, as adversaries with access to GPUs or botnets can solve them cheaply at scale (R1)~\cite{abadi2005moderately, green2011reconstructing, ali2020foundations}. Deployment of PoW puzzles does not introduce immediately new attack vectors to compromise the server (R2).
Finally, PoW can be tuned to match heterogeneous agent capabilities (C1), but are not suitable for agents that cannot run specialized PoW software (C2)~\cite{guerra2022design}.

\newlength{\diffcolwd}
\setlength{\diffcolwd}{2cm} 

\begin{table*}[!t]
  \centering
  \small
  \caption{\textbf{Representative examples of rRGs across difficulty levels.} Each problem requires solving interdisciplinary clues (e.g., humanities, social sciences, mathematics, or sciences), extracting the first letters of the answers, and combining them to form a hidden word. As difficulty increases, the number of clues grow, domains become more diverse, and reasoning steps deepen, illustrating how our framework controls challenge complexity.}
  \label{tab:difficulty-prompts}
  \begin{tabular}{@{}p{\diffcolwd}|p{\dimexpr\textwidth-\diffcolwd\relax}@{}}
    \toprule
    \textbf{Difficulty} & \textbf{Example Problem} \\
    \midrule
    Easy &
      \textit{Begin your journey by deciphering these interdisciplinary clues. For each clue, determine the answer and record the first letter. The sequentially extracted letters will reveal the hidden word.}\newline
      • Humanities: Ancient epic that recounts the siege of Troy.\newline
      • Social Science: Term describing the layering of social classes.\newline
      • Mathematics: Simplest geometric figure, a straight unbroken path.\newline
      • Social Science: Immersive qualitative research method used by anthropologists.\newline
      • Social Science: Discipline devoted to studying human society.\newline
      \textbf{Solution:} \texttt{isles} \\
    \midrule
    Medium &
      \textit{Decode the enigma by solving each interdisciplinary clue below. For each, determine the appropriate answer and extract its initial letter; in sequence, they will reveal the hidden word.}\newline
      • Computer Science: Central component that directs operations in a computer system.\newline
      • Social Science: A societal process in which a community absorbs external influences.\newline
      • Mathematics: The operation that produces the least upper bound in lattice theory.\newline
      • Physics: A phenomenon in wave mechanics where motion reaches peak amplitude.\newline
      • Humanities: Art movement emphasizing emotion and individualism.\newline
      • … (5 more clues omitted for brevity)\newline
      \textbf{Solution:} \texttt{majordomos} \\
    \midrule
    Hard &
      \textit{Embark on the following multifaceted challenge by deciphering each clue. For each interdisciplinary riddle, determine the answer and record its initial letter. When read in order, these letters will spell out the hidden word.}\newline
      • Astronomy: Term for matter found between star systems.\newline
      • Chemistry: Analytical procedure using gradual chemical addition.\newline
      • Social Science: Concept reflecting fairness and equal opportunity.\newline
      • Physics: Energy transfer phenomenon at a specific frequency.\newline
      • Earth Science: Study of crustal shifts in tectonics.\newline
      • … (6 more clues omitted for brevity)\newline
      \textbf{Solution:} \texttt{interverting} \\
    \bottomrule
  \end{tabular}
\end{table*}

\subsection{Towards Reasoning Gates}
As discussed in Section~\ref{sec:existing_approaches}, existing throttling approaches fail to achieve the desired computational imbalance: they are either too easy for modern agents to bypass using increasingly powerful language models, or they rely on VLMs and thus assume agents can process images, which excludes many text-only agents. To impose meaningful overhead, we need puzzles that are inherently more challenging and cannot be solved through simple tricks or shortcuts.  Since the targeted agents themselves rely on LMs, the natural direction is to design puzzles that force them into costly reasoning. Multi-step inference and symbolic composition require generating many additional tokens, which directly increases both the latency and financial cost of solving each challenge.  

We propose \emph{reasoning gates} as a promising solution. Reasoning gates are throttling gates instantiated with reasoning-based puzzles that demand multi-hop inference and symbolic composition, tasks that remain challenging even for state-of-the-art models. By leveraging the costliness of reasoning token generation, reasoning gates address the shortcomings of existing throttling mechanisms and naturally create the desired computational asymmetry. The remaining question, then, is how to identify and construct suitable reasoning problem sets for use within these gates, which we discuss next in Section~\ref{sec:existing_prob_sets}.

\subsection{Existing Reasoning Problem Sets}\label{sec:existing_prob_sets}
To develop effective throttling mechanisms for LM-based web agents, we assess whether existing challenge formats satisfy the key properties outlined in Section~\ref{sec:properties}. While logic, coding, and scientific benchmark test reasoning, they often fail to resist solver shortcuts, scale dynamically, or support real-world deployment. This section analyzes these limitations against existing benchmarks and motivates the need for a new scalable and robust challenge generation framework:

\pbe{Solver-based Problems.} Solver-based tasks, such as logic puzzles and symbolic reasoning~\cite{pan2023logic, feng2023language} can be used as challenges to the agent. However, generating these problems is cheap (A1), e.g., requires only sampling ground truth parameters, and verification of correctness is straightforward (A2) while solution is expensive (A3)~\cite{zebralogic}. For scalability (S), some diversity can be achieved through logic templates (S1), and operational overhead to generate and verify these puzzles by the server is low (S2). However, solver-based problems are fragile: adversaries can “cut corners” by integrating off-the-shelf solvers (R1), though these puzzles do not introduce significant new server-side attack surfaces (R2). Finally, such problems can be tuned in difficulty by adjusting logical depth (C1), and they are compatible with text-only and low-resource agents (C2).

\begin{figure*}[!t]       
  \centering
  \includegraphics[width=\textwidth]{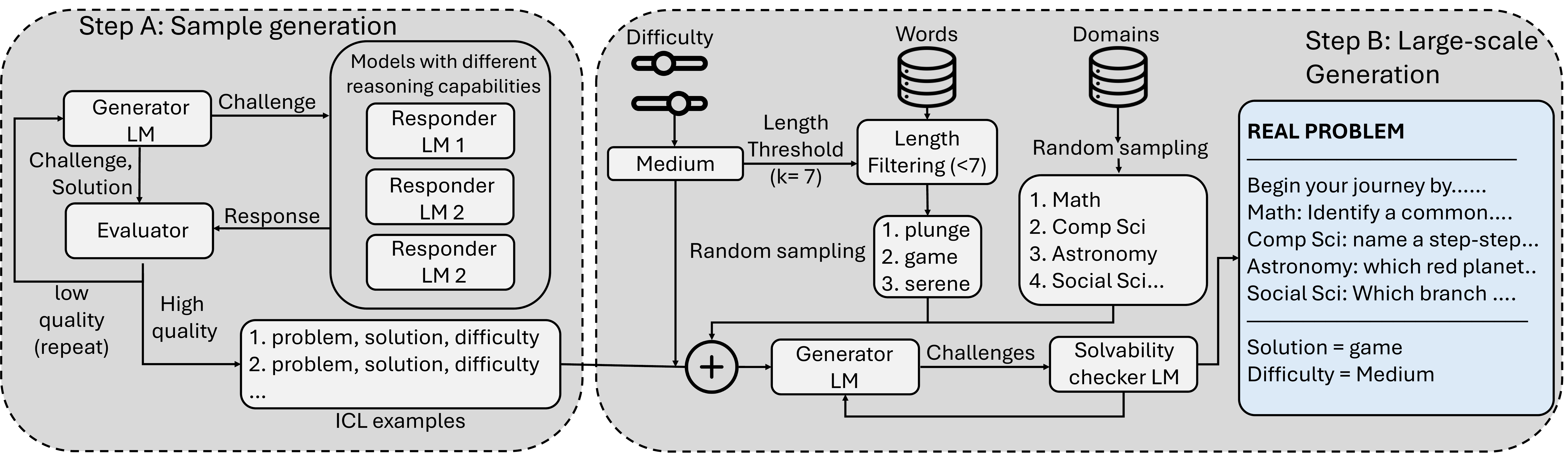}
  \caption{\textbf{Reasoning gate generation pipeline.} The framework pre-computes rRGs in two steps. \textit{Step A (Sample Generation):} A generator LM produces candidate challenges, which are filtered by evaluators and responder models of varying reasoning capabilities to construct high-quality ICL example labeled with difficulty. \textit{Step B (Large-scale generation):} Using these examples, the generator LM creates a large challenge bank by sampling words and domains, applying length filtering, and verifying solvability. Each challenge is labeled with a solution and difficulty, producing scalable puzzles for throttling  agents.}
  \label{fig:offline_generation}
\end{figure*}

\pbe{Coding-based Problems.} Programming challenges introduce diversity and open-endedness, but they exhibit several limitations when evaluated against throttling properties. Generating tasks is expensive because it requires curating and mutating functional code snippets at scale (A1), and verification is costly since it demands sandboxed execution frameworks that guard against malicious behavior (A2)~\cite{alhindi2025threadbox}. Solving coding problems will result in high token count, making solution expensive (A3).
With respect to scalability (S), coding problems can be created in many variations (S1), but reliably generating solvable, diverse tasks without overlap is difficult and verification overhead scales poorly (S2). On robustness (R), adversaries can exploit tool integration and code memorization~\cite{chen2025memorize} to shortcut solving (R1), while sandboxing itself introduces attack surfaces (R2) since verification frameworks must execute untrusted code. Finally, coding problems can be adjusted in difficulty (C1) by changing task complexity, and text-only and low-resource agents can still generate solution code (C2)~\cite{sepidband2025enhancing, wang2025epicoder, bogin2024super}.

\pbe{Advanced Sciences Problems.} Scientific and mathematical challenges~\cite{gao2024omni, sun2025challenging, gulati2024putnam, he2024olympiadbench} demonstrate strong reasoning requirements but remain impractical as throttling mechanisms. Generation is extremely costly because problems must be hand-curated by domain experts (A1), verification is non-trivial and often requires detailed grading rubrics (A2), and while solving is expensive for LMs (A3), the asymmetry is undermined by their static, limited pool. In addition, diversity is limited (S1) and operational costs are prohibitive as it requires contributions from experts across different fields (S2)~\cite{phan2025humanity}. In terms of robustness (R), they are more resistant to shortcuts (R1) since no simple off-the-shelf solver can easily bypass them, but their static nature makes them vulnerable to memorization and fine-tuning exploits which is an open research problem~\cite{huang2025math, yu2025benchmarking, yao2025var, xie2024memorization}, while verification complexity risks new attack surfaces for the server (R2). Finally, difficulty can be tuned by selecting between undergraduate- and Olympiad-level problems (C1), and could be made accessible for text-only and low-resource agents (C2).

\subsection{rRG: Rebus-Based Reasoning Gates}
\label{sec:rebus_intro}
As discussed before, conventional reasoning challenges (e.g., math and coding) suffer from these limitations because they are constraint-heavy and have a larger set of rules, making it difficult to satisfy the strict formal requirements. As a result, generation typically requires manual curation to ensure constraints aren’t violated, and the tasks are often less robust given the availability of classical solution methods. We find soft reasoning challenges do not suffer from the same disadvantages.

\pbe{Soft Reasoning Challenges.}
Instead of relying on scientific or mathematical challenges, recent benchmarks also assess a language model’s ability to infer implicit details from a described world or environment. These challenges are called soft-reasoning challenges because the problems may not directly state the rules or restrictions that govern the world~\cite{zhang2025soft,sprague2023musr}. Although they are scalable (LMs can synthesize fictional settings and leverage existing literature),prior work suggests that LM's with varying capabilities do not perform differently on these reasoning tasks~\cite{ahuja2025finding}, making reliable difficulty control challenging.

\pbe{Rebus-based Challenges.} To achieve scalability comparable to soft-reasoning tasks while retaining the difficulty control of conventional reasoning, we use an idea from rebuses~\cite{gritsevskiy2024rebus}, where the task is to guess a word where every letter is constructed through a visual puzzle,  and design textual challenges---rebus-based reasoning gates (rRG). We pick a random word and for each letter we generate a question about a fact from a diverse set of domains, e.g., computer science, psychology etc. Since the clues are domain-based, the model's performance should improve with increased reasoning~\cite{rein2024gpqa, openaiIntroducingOpenAI}.
Individual clues require no manual validation and rely only on factual or conceptual knowledge, yet they can be composed into multi-hop reasoning challenges~\cite{wei2025satbench}, turning simple facts into a structured inference task. Because text-rebuses follow simple rules—assemble an English word from domain-specific clues—stronger reasoning models can better spot inconsistencies and, with backtracking~\cite{muennighoff2025s1}, iteratively correct errors, that could improve performance.

Since we aim to generate text-based multi-hop reasoning challenges with each hint rooted in factual clues LMs become a strong and feasible candidate as they are trained on vast amounts of information from different domains. Further, LMs have been shown to have strong capabilities to generate challenges with different reasoning requirements and complexities~\cite{patel2024multi}. Table~\ref{tab:difficulty-prompts} shows examples of rRGs.

\section{Reasoning Gates Framework}
\label{sec:reasoning_gates}

To make the framework feasible for the service provider $\mathcal{S}$, challenges issued by $\mathcal{S}$ and generated via algorithm $\mathcal{G}$ are produced offline, so that the latency is minimal during real-time interaction between $\mathcal{S}$ and $\mathcal{A}$. As illustrated in Figure~\ref{fig:offline_generation}, $\mathcal{G}$ builds the offline generated challenge bank $C_b$ in two steps. Step A constructs a set of \textit{high-quality} examples $C_e$ to be used as in-context learning (ICL) examples for large-scale generation via many-shot learning~\cite{agarwal2024many}. These examples are created through interactions with multiple models across difficulty levels, yielding examples that capture the desired complexity for each level (higher difficulty implies greater required reasoning ability). Step B uses the $C_e$ to generate the bank $C_b$, that $\mathcal{S}$ will sample from during the online phase, populating $C_b$ with challenges at varying difficulty levels using a large set of words and domains as input.

The generation can be formulated as:
\begin{align*}
\begin{split}
C_b = \mathcal{G}([d_1,...,d_m],[w_1,...,w_m])
\end{split}
\end{align*}
where $[d_1,...,d_m]$ is the bank of knowledge domains or subject matters and $[w_1,...,w_m]$.

$\mathcal{S}$, as illustrated in Figure~\ref{fig:main}, uses algorithm $\mathcal{T}$ to throttle agents by sampling batch of challenge-solution pairs ($c,z$) from $C_b$ for each round, until the rejection or access criteria is met.

The $\mathcal{S}-\mathcal{A}$ interaction phase can be formulated as:
\begin{align*}
\begin{split}
\text{Access Permission} = \mathcal{T}(C_b, l, T_{num}, T_{min})
\end{split}
\end{align*}
where $T_{num}$ is the total number of challenges to be issued and $T_{min}$ is the minimum number of challenges $\mathcal{A}$ needs to solve to get access.

The following subsections give a detailed description of Algorithm $\mathcal{G}$ (offline phase) and Algorithm $\mathcal{T}$ (online phase).

\subsection{Precomputed Reasoning Gates (Offline)}
\label{sec:offline_phase}
Algorithm~\ref{alg:stepA} outlines step A of Algorithm $\mathcal{G}$, the generation of the ICL challenge bank $C_e$. It selects a set of response models ($\mathcal{A}^t$), each with varying reasoning capabilities, and uses them as proxies to estimate the complexity of the challenges.

For each desired difficulty level, in set of all possible difficulty levels ($\mathcal{L}$), only a subset of $\mathcal{A}^t$ should be able to solve the challenge, e.g. if the difficulty is set to low then all the models should be able to solve the challenge and if difficulty is set to high then only the models with the strongest reasoning abilities should be able to solve the challenge. To populate $C_e$ the algorithm creates a challenge bank $C_\ell$ for each difficulty level $\ell \in \mathcal{L}$, and the union of these banks forms $C_e$.
Challenges in $C_\ell$ are generated by using a sample generation LM $G_{icl}$ that utilizes a system prompt (Appendix~\ref{appendix:verifier_prompt}) written to support tuning the difficulty(Appendix~\ref{appendix:difficulty_prompt}) of the next challenge based on challenge history. 
First, the algorithm collects the responses of each model in $\mathcal{A}^t$. If challenge is solved by the models in $\mathcal{A}^t_\ell$ exclusively, and no other model in $\mathcal{A}^t$ can solve it, then the algorithm adds $(c,z)$ to $C_\ell$ followed by sending a request to $G_{icl}$ to generate a challenge of the same difficulty.

If more models than the ones present in $\mathcal{A}^t_\ell$ solve the challenge then the next requested challenge is of increases difficulty, as it is easier than desired. If models in  $\mathcal{A}^t_\ell$ are not able to solve it then an easier challenge is requested. Besides the seed challenge, $G_{icl}$ utilizes the generation history as context alongside the desired change in difficulty level for the generation of the subsequent challenge. This process terminates when $C_e$ contains the required number of challenges for all difficulty levels.

~

\begin{algorithm}[H]
\caption{Step A — Prepare ICL Examples for Each Difficulty Level}
\label{alg:stepA}
\begin{algorithmic}[1]
\Require ICL challenge generation LM $G_{icl}$, Solvability checker LM $S$, Response generation LMs $\mathcal{A}^t = \{A^t_1,\dots,A^t_m\}$, Desired difficulty levels $\mathcal{L}$, Number of ICL examples per difficulty $k$
\Ensure ICL example bank $C_e$ labeled by difficulty
\State Initialize empty example challenge list $C_e$
\ForAll{$\ell \in \mathcal{L}$}
  \State Initialize empty challenge example list $C_\ell$
   \PreCmt{Desired model subset per selected $\ell$}
  \State Select $\mathcal{A}^t_\ell \subseteq \mathcal{A}^t$
  \PreCmt{Seed challenge and solution generation}
  \State $c, z \gets \textsc{Challenge-Gen}(\varnothing, \text{None})$ 
  \While{$|c_\ell| < k$}
    \State $R_\ell \gets \big[\,\mathcal{A}(c)\ \forall\ \mathcal{A} \in \mathcal{A}^t_\ell\,\big]$
    \State $R_c \gets \big[\,\mathcal{A}(c)\ \forall\ \mathcal{A} \in \mathcal{A}^t \setminus \mathcal{A}^t_\ell\,\big]$
    \If{$\text{set}(z) = \text{set}(R_\ell)$ }
    \If {$\text{set}(R_\ell) \cap \text{set}(R_c)$ is $\varnothing$} 
        \State Append $(c,z)$ to $C_\ell$
        \State $tuning\_cmd \gets \text{Maintain Difficulty}$
    \Else
        \State $tuning\_cmd \gets \text{Increase Difficulty}$
    \EndIf
    \Else
        \State $tuning\_cmd \gets \text{Decrease Difficulty}$
    \EndIf
    \State $c, z \gets \textsc{Challenge-Gen}(C_\ell,\ tuning\_cmd)$
  \EndWhile
  \State Append $(C_\ell, \ell)$ to $C_e$
\EndFor
\State \Return $C_e$
\PreCmtLong {previous generation history $C_h$ and difficulty adjustment command (maintain, increase, decrease)}
\Function {Challenge-Gen}{$C_h$, tuning}

    \While{not terminated}
        \If{$C_h = \varnothing$ \textbf{ and } $tuning=\text{None}$}
            \State $c,z \gets G(\varnothing,\ \text{First Problem})$
        \Else
            \State $c,z \gets G(C_h,\ tuning)$
        \EndIf
        \If{$S(c,z)=\text{True}$}
            \State \Return $c,z$
        \EndIf
    \EndWhile
\EndFunction
\end{algorithmic}
\end{algorithm}

\begin{algorithm}
\caption{Step B — Large-Scale Challenge Generation}
\label{alg:stepB}
\begin{algorithmic}[1]
\Require Large-Scale challenge generation LM $G$, Solvability checker LM $S$, domains $D$, words $W$, difficulty levels $\mathcal{L}$, target problem count $N$, ICL example bank $C_e$
\Ensure Offline challenge bank $C_b$ with labeled difficulties
\State Initialize $C_b \gets \emptyset$
\While{$|C_b| < N$}
  \State Sample target difficulty $\ell \in \mathcal{L}$
  \State Sample domains $d_1,\dots,d_m$ from $D$
  \State Sample words $w_1,\dots,w_m$ from $W$
  \State $c, z \gets G(d_1,\dots,d_m,\ w_1,\dots,w_m,\ C_e,\ \ell)$
  \If{$S(c,z)=\text{True}$}
      \State Append $(\ell, c, z, d_1,\dots,d_m,\ w_1,\dots,w_m)$ to $C_b$
  \EndIf
\EndWhile
\State \Return $C_b$
\end{algorithmic}
\end{algorithm}

Algorithm~\ref{alg:stepB} describes step 2 of Algorithm $\mathcal{G}$, the large-scale generation process. It employs a generation model ($G$), a solvability checker ($SC$), $C_e$ for many-shot prompting, and a large set of diverse domains ($D$) and words ($W$). The procedure begins with an empty challenge bank $C_b$, which is then populated with the desired number of challenges at each difficulty level. For each challenge, $G$ is provided with randomly sampled words and domains (e.g., Classics, Entomology, Environmental Toxicology) sampled from $W$ and $D$, together with $C_e$ as input. Random sampling from predefined banks reduces the likelihood of repetition: unlike unrestricted language model generation, it mitigates biases introduced by $G$’s inherent word and domain preferences, leading to a more diverse set of challenges. Before the algorithm adds the challenge to the bank, it verifies its correctness using $SC$.
\subsection{Interactive Verification (Online)}
\label{sec:online_phase}
Algorithm~\ref{alg:online} describes Algorithm $\mathcal{T}$ used for interactive verification phase between the service provider $\mathcal{S}$ and the agent $\mathcal{A}$. In Algorithm $\mathcal{T}$, $\mathcal{S}$ sets $T_{max}$ as the desired number of maximum challenges per interaction between $\mathcal{S}$ and $\mathcal{A}$. $T_{min}$ is the minimum number of challenges $\mathcal{A}$ needs to answer correctly. $\mathcal{S}$ starts by sending a challenge $c$ of desired difficulty $\ell$, randomly sampled from $C_b$, to $\mathcal{A}$ and waits for the response string $r$. Once it receives $r$ it compares it to the solution string $z$. If $r$ and $z$ are equal, it counts it towards the correct responses by $\mathcal{A}$, otherwise it is counted towards incorrect responses. 

This challenge-response session continues until the time $T_{max}$ is reached. If $\mathcal{A}$ solves more than or equal to $T_{min}$ challenges then it is granted access, it is denied access otherwise.

\subsection{Evaluating the  Desired Properties}
\label{sec:satisfaction_properties}
This section explains how our framework satisfies the desired properties discussed in Section~\ref{sec:properties}. 

\pbe{Computational Asymmetry.} Gates are generated with smaller models (e.g., o3-mini or GPT-4o) rather than state-of-the-art models such as OpenAI o3 or Anthropic Claude 3.7, or with reasoning models under strict reasoning-token budgets (e.g., Gemini 2.5 with $\leq 100$ tokens). We evaluate different models in Section~\ref{sec:exps}. Moreover, even when the same model is used as both generator and responder, generation consumes fewer tokens than solving. We hypothesize that drawing on randomly sampled words, domains, and patterns shrinks the generator’s search space, whereas responding requires multi-step reasoning over a larger space. This design satisfies \textbf{A1} and \textbf{A3}. Since each solution is a single string, we verify in constant time by precomputing solution hashes offline and checking equality online, satisfying \textbf{A2}.
~\begin{algorithm}
\caption{Online Phase: Interactive Verification}
\label{alg:online}
\begin{algorithmic}[1]
\Require Offline problem set $C_b$, Desired difficulty level ($\mathcal{L}$), Min correct responses $T_{min}$, Max num of challenges $T_{num}$, prover (agent) $\mathcal{A}$
\Ensure Access permission
\State Initialize Access Permission $\!\leftarrow\!$ False
\State Initialize total challenges sent $C_s \!\leftarrow\!0$
\State Initial total correct number of responses $C_r \!\leftarrow\!0$
\While{$C_s$ less than $T_{num}$}
  \State Select random (\textit{c, z}) from $C_b$ with difficulty $\mathcal{L}$
  \State Send \textit{c} to $A$, await \textit{r}
  \If{$z$ equal $r$}
    \State $r \gets C_r + 1$
  \EndIf
  \State $C_s \gets C_s + 1$
\EndWhile
\If{$C_r$ is greater than or equal to $T_{min}$}
\State Access Permission $\!\leftarrow\!$ True
\EndIf
\State \Return Access Permission
\end{algorithmic}
\end{algorithm}

\pbe{Scalability.} The generation pipeline shown in Figure~\ref{fig:offline_generation} allows for automated large-scale challenge generation in an offline phase, satisfying condition \textbf{S1}. The generation can be done using existing LM APIs for cheap and the challenges are stored as strings, making the storage and retrieval of the challenges simple and lightweight. The online challenge-response phase relies on retrieval of text-based challenges and string matching, avoiding bottlenecks in the online phase and satisfying property \textbf{S2}.

\pbe{Robustness to Adversaries.}
Because of the challenges’ complexity, longer target words, domain diversity, time limits, and symbolic-reasoning requirements, these tasks are difficult for gig workers or off-the-shelf solvers (see Table~\ref{tab:human_eval}). An adversary cannot reduce cost by intentionally answering incorrectly; wrong answers simply deny access, enforcing the cost of a successful bypass and satisfying \textbf{R1}. Since each solution is a single string, verification reduces to straightforward string matching, avoiding third-party verification frameworks that may introduce vulnerabilities, thus satisfying \textbf{R2}.

\pbe{Compatibility.} 
The reasoning gates framework generates problems with different difficulty levels, $\mathcal{S}$ can adjust the difficulty based on desired throttling goal. Our evaluation in Section~\ref{offiline_difficulty} shows that the framework is successful in generating challenges of different difficulty levels, satisfying properties \textbf{C1}. Since the entire framework generates text-rebuses, hence providing textual challenges and expecting a word (string) as a response, the framework supports text-only web agents satisfying property \textbf{C2}.

\subsection{Deployment of Reasoning Gates}
\label{sec:deployment}

\pbe{Web Interface.} Since web agents are capable of navigating web services through user interfaces, one way of implementing Reasoning Gates is through directly prompting the agent through the interface, like CAPTCHA, to solve the challenge before accessing the resource. This makes the process simpler and doesn't require the agent to support additional tools for resource access but can incur an additional delay in solving the challenge.

\pbe{MCP, API-Only Servers, and Background Processes.}
Reasoning Gate can be implemented to secure web-service API endpoints using MCP~\cite{anthropicIntroducingModel}. The service can present a challenge prior to content delivery and, upon success, issue a token that authorizes subsequent access for the remainder of the session. Alternatively, the challenge may be dispatched to the agent for background solving, similar to Cloudflare’s Turnstile~\cite{cloudflareEmbedWidget}.

\pbe{Cascading with Authentication and Attestation.}
Reasoning Gates can be implemented alongside attestation services to improve user accessibility, especially for the users who do not have access to agents and are running legacy systems~\cite{beekman2016attestation, cloudflareHumanityWastes}. The user support can also be improved through integration of user authentication services such as Multi-Factor-Authentication (MFA)~\cite{ometov2018multi}, allowing them to bypass reasoning gates. Integrating reasoning gates with other authentication mechanisms can also reduce the number of challenges required to get access and also allows the service provider to trust the user, making the service access easier.

\pbe{Benign Users and Web Agents.} 
We assume that a benign user or web agent can support at least one of the following capabilities. For legacy users without agents, authentication and attestation grant access without requiring them to solve agent-oriented challenges. For web agents, requests fall into two categories: (1) interface-based navigation, handled by the framework’s web-UI support, and (2) API access, handled via the framework’s MCP integration.

\begin{figure*}[t]
  \centering
  \begin{subfigure}{0.3\linewidth}
    \centering
    \includegraphics[width=\linewidth]{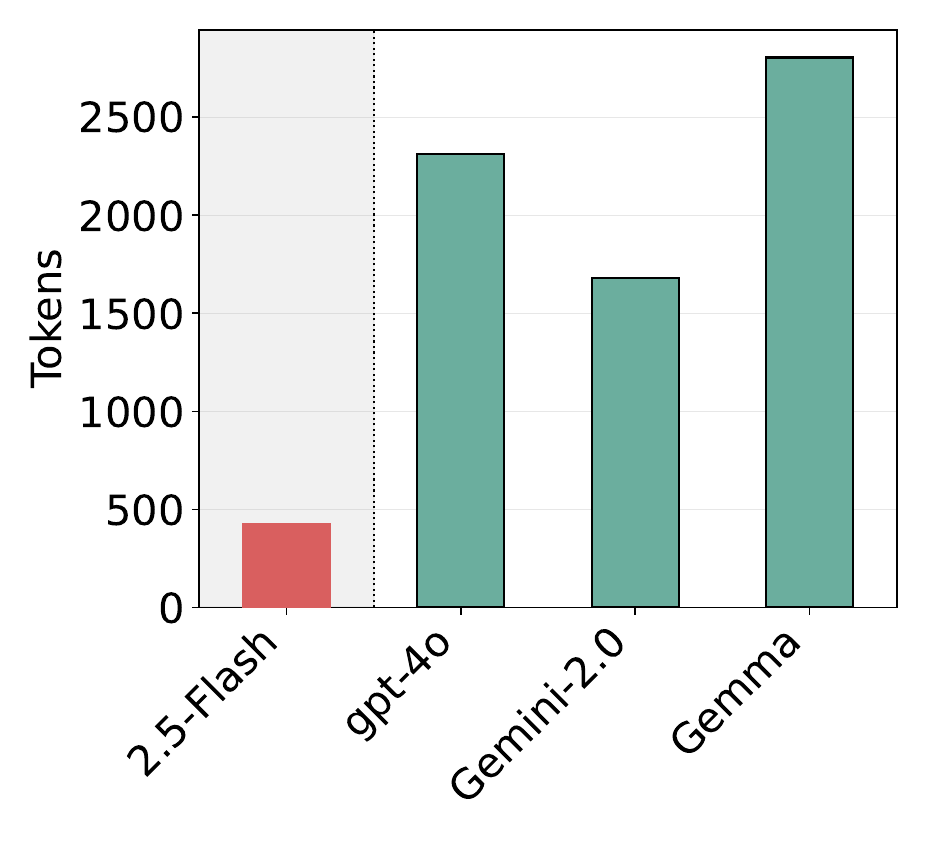}
  \end{subfigure}
      \hspace{-0.1in}
  \begin{subfigure}{0.3\linewidth}
    \centering
    \includegraphics[width=\linewidth]{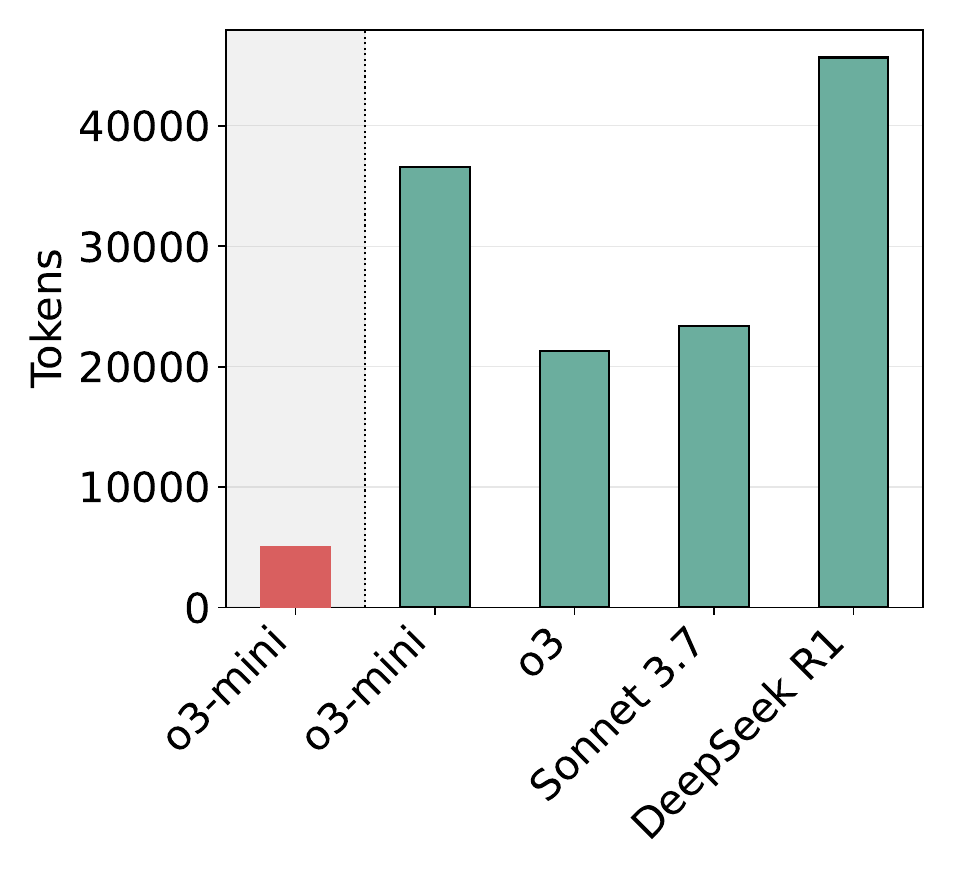}
  \end{subfigure}
  \caption{%
    \textbf{Comparison of token efficiency between generation and prover models.}  
    (left) Average tokens used by the generation model vs.\ total tokens used by the prover model (no reasoning).  
    (right) Average tokens used by the generation model vs.\ total tokens used by the prover model (with reasoning).
  }
  \label{fig:token_efficiency_comparison}
\end{figure*}

\section{Experiments}
\label{sec:exps}

To evaluate our rebus-based Reasoning Gates against the criteria in Section~\ref{sec:properties}, we begin by examining \textit{asymmetry}, comparing the token and compute cost of challenge generation with that of response generation in a single interaction session. We then assess model performance and analyze \textit{scalability} by measuring how model choice affects the complexity and quality of generated problems, followed by quantifying hallucination rates and the incidence of unsolvable items during generation. Next, we conduct a user study to test whether a \textit{gig-worker shortcut} can bypass the system. Finally, we validate \textit{difficulty control} by testing models with different reasoning capabilities and by comparing against other reasoning benchmarks.


\subsection{Experimental Setup}

\paragraphbe{Models.} We evaluate our framework using multiple reasoning (OpenAI o3 and o3-mini, DeepSeek R1, Claude 3.7 Sonnet) and non-reasoning models (OpenAI GPT-4o and Google Gemma 3-27B, Claude 3.5 and Gemini 2.0-Flash). We utilize o3-mini and Gemini 2.5-Flash, with a thinking budget of 500 tokens, as our primary challenge generator models. 

\paragraphbe{Dataset.} 
We assess the performance of our framework's challenge generation capabilities by comparing our generated challenges with MuSR~\cite{sprague2023musr} and FlawedFiction~\cite{ahuja2025finding} benchmark, both designed to assess multi-step soft reasoning in narrative contexts. 

We use two reasoning tasks from MuSR dataset: \textit{Murder Mysteries} (250 samples) and \textit{Object Placement} (250). Murder Mystery requires the LM to find the criminal from the list of suspects. These challenges assess the models capability to reason through social concepts and understand motives. Object Placement requires the LM to determine the position of a requested object, testing its spatial reasoning skills. FlawedFiction assesses models reasoning capabilities through story comprehension. This benchmark takes existing stories and rewrites them to introduce a single continuity error; each story sample contains exactly one error.

\paragraphbe{Evaluation Metrics.} We use accuracy on generated samples and existing benchmarks to assess the efficacy of the framework in generating challenges of different complexities. 
For the cost metric we count number of generated output tokens. We use it to measure the cost of generated challenges and the cost incurred on the agent to respond to them. 
We choose tokens over monetary cost as cost per millions token do not reflect the actual cost of generation of challenges. It is often shaped by service provider's commercial priorities such as market acquisition strategies. Furthermore, using output tokens also makes it easier to compare open-source and closed-source models. 

\subsection{Results and Analysis} 

\paragraphbe{Throttling Performance and Asymmetry Analysis.}
To analyze the throttling mechanism (see Figure~\ref{fig:token_efficiency_comparison}), we generate challenges with two models: Gemini 2.5-Flash for non-reasoning models—using a 100-token thinking budget—and o3-mini for reasoning models (see Figure~\ref{fig:token_efficiency_comparison}). For both reasoning and non-reasoning models, we pick the challenges from the generators' \textit{extremely difficult} challenge set. This allows us to observe the maximum achievable throttling of the prover models.
\begin{figure}[!t]
  \centering
  \includegraphics[width=0.9\columnwidth]{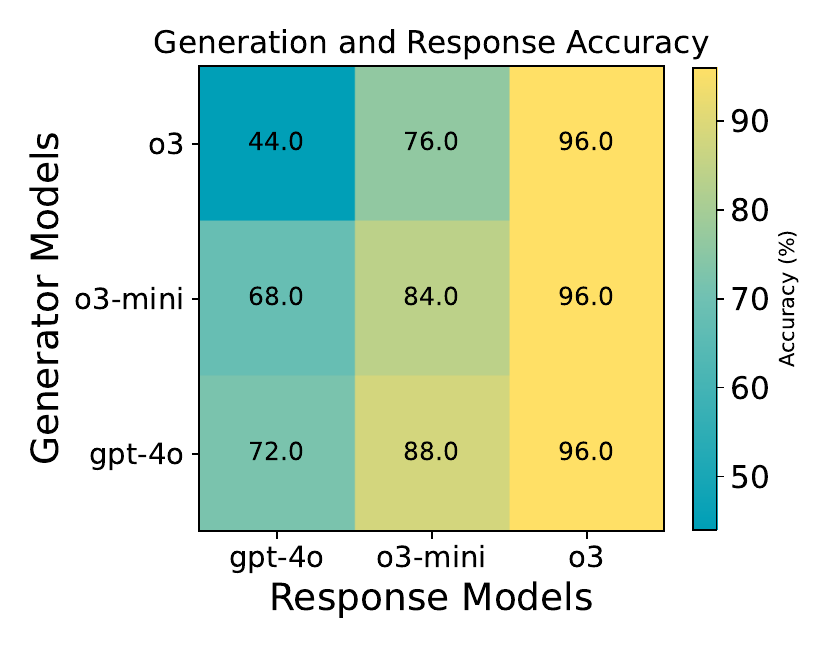}
  \caption{\textbf{Accuracy of models when used as both challenge generators (rows) and responders (columns)}. Stronger generator models (e.g., o3) produce harder and more discriminative challenges, leading to sharper accuracy drops for weaker responders, while o3 remains robust across all generated tasks.}
  \label{fig:generation_heatmap}
\end{figure}
For non-reasoning models, Gemini 2.5 takes 430 tokens for challenge generation and Gemma (the highest token consuming prover model) utilizes 2.8k tokens to reach a correct response, achieving a computational asymmetry of \textbf{6.6x}. On average, we achieve computation asymmetry of \textbf{5.2x} for non-reasoning models. For reasoning models, o3-mini takes approximately 4.9k tokens for challenge generation and DeepSeek R1 (the highest token consuming prover model) utilizes 46k tokens to reach the correct response, achieving an asymmetry of \textbf{9.2x}. On average we achieve an asymmetry of \textbf{6.4x} for reasoning models.

These results show that, for both reasoning and non-reasoning models, the generation cost is significantly cheaper than the verification cost by significant margin, achieving low challenge generation cost (property \textbf{A1}) and high response generation cost (property \textbf{A3}). 

\begin{table}[ht]
\centering
\small
\renewcommand{\arraystretch}{0.9}  
\caption{\textbf{Comparison of success rates and completion times between humans, Browser-Use agents (Web Interface), and MCP-based agents.} For the MCP setup, we used FastMCP as the server implementation and the Agent Development Kit (ADK) as the agent framework. Human participants solved only a subset of challenges—even with Internet access—whereas agents consistently achieved perfect success with substantially lower latency.}
\label{tab:human_eval}
\resizebox{\columnwidth}{!}{
\begin{tabular}{llcc}
\toprule
\textbf{Category} & \textbf{Condition / Model} & \textbf{\shortstack{Access \\ Granted}} & \textbf{\shortstack{Avg. Time \\ (mm:ss)}} \\
\midrule
\multirow{2}{*}{Humans} 
    & w/o Internet & 2/10  & 11:30 \\
    & w/ Internet  & 6/10  & 11:00 \\
\midrule
\multirow{2}{*}{Web Interface} 
    & o3-mini      & 10/10 & 2:35  \\
    & o3           & 10/10 & 4:50  \\
\midrule
\multirow{2}{*}{MCP} 
    & o3-mini      & 10/10 & 0:30  \\
    & o3           & 10/10 & 0:17  \\
\bottomrule
\end{tabular}}
\end{table}

\paragraphbe{Impact of Model Selection.} In this section, we analyze language models with varying reasoning capabilities as both generators and responders. Figure~\ref{fig:generation_heatmap} compares GPT-4o, o3-mini, and o3 in both roles. The heatmap shows that stronger generator reasoning produces challenges that better discriminate among responder capabilities: for example, while all generators yield similar accuracy on o3 (the most capable model considered), accuracy drops more steeply as responder reasoning decreases. This indicates that higher-reasoning generators produce more complex and diverse challenges that affect models of different reasoning capabilities differently.

\begin{figure}[ht]
\centering
\includegraphics[width=0.5\columnwidth]{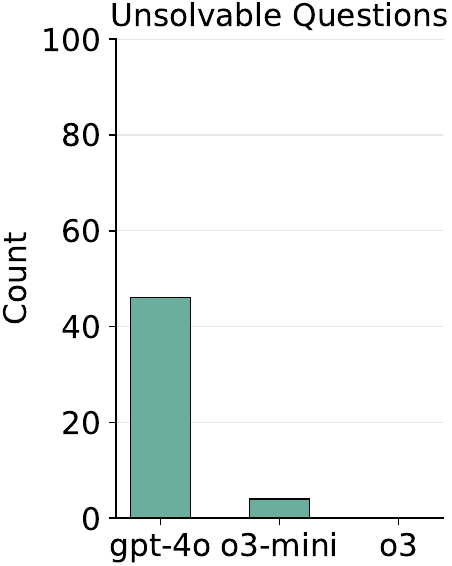}
\caption{\textbf{Number of unsolvable questions per 100 generated questions.}}%
\label{fig:unsolvable_questions}
\end{figure}

Figure~\ref{fig:unsolvable_questions} reports the number of unsolvable questions produced by each generator. Per 100 generated items, GPT-4o produces 46 (highest), o3-mini produces 4, and o3 produces 0 (lowest). Thus, greater generator reasoning capability yields fewer unsolvable challenges.

\begin{figure}[ht]
\centering
\includegraphics[width=0.85\columnwidth]{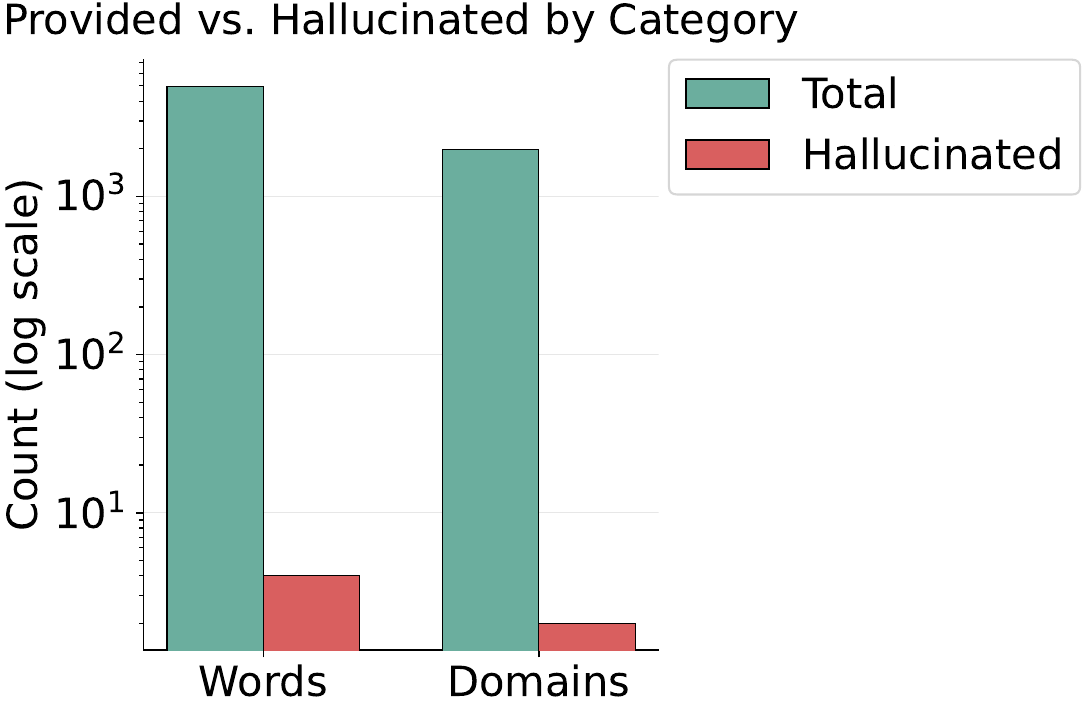}
\caption{\textbf{Hallucination rate of generation.} Rate of words and domains that were generated outside of the given lists.}%
\label{fig:total_hallucinated}
\end{figure}

Since the large-scale generation takes place using random sampling words and domains, it is important that the generator samples only from the provided words and domains list does minimal hallucination. In Figure~\ref{fig:total_hallucinated}, we show the number of words and domains used by the generator model that were not the part of provided words and domains bank. For over 4000 words utilized only 4 words were hallucinated and for close to 2000 domains utilized only 2 were hallucinated, making the approximate hallucination rate 0.01\%. 
These results indicate that using reasoning models (e.g., o3 and o3-mini) in conjunction with random sampling over domains enables diverse and scalable generation while minimizing hallucination of words and domains (property \textbf{S1}).

\begin{table}[h]
\centering
\small
\caption{\textbf{Step A Difficulty.} Accuracy (\textbf{A}, \%) and output tokens (\textbf{OT}, 1k) of GPT-4o, o3-mini, and o3 on offline Step A examples, showing higher performance for stronger reasoning models and a clear drop as difficulty increases.}
\label{tab:icl_difficulty}
\setlength{\tabcolsep}{8pt}
\begin{tabular}{@{}l c @{\hspace{0.1cm}}c  c@{\hspace{0.1cm}}c c@{\hspace{0.1cm}}c @{}}
\toprule
\textbf{Model} &
\multicolumn{2}{c}{\textbf{Easy}} &
\multicolumn{2}{c}{\textbf{Medium}} &
\multicolumn{2}{c}{\textbf{Hard}} \\
\cmidrule(lr){2-3} \cmidrule(lr){4-5} \cmidrule(lr){6-7}
& \textbf{A(\%)  $\downarrow$ } & \textbf{OT  $\uparrow$} & \textbf{A(\%)  $\downarrow$ } & \textbf{OT  $\uparrow$} & \textbf{A(\%)  $\downarrow$ } & \textbf{OT  $\uparrow$} \\
\midrule
GPT-4o   & 75.0  & $0.48$   & 22.2  & $0.46$   & 0.0   & $0.32 $ \\
o3-mini  & 84.0  & $6.25$ & 77.8  & $6.41$ & 0.0   & $6.76$ \\
o3       & 87.5  & $4.43$ & 88.9  & $6.74$ & 30.1  & $5.76$ \\
\midrule
\textbf{Average} & \textbf{82.2} & \textbf{$3.72$} & \textbf{63.0} & \textbf{$4.54$} & \textbf{10.0} & \textbf{$4.28$} \\
\bottomrule
\end{tabular}
\end{table}

\begin{table}[ht]
  \centering
  \small
  \renewcommand{\arraystretch}{0.9}
  \caption{\textbf{Reasoning Gates comparison.} Accuracy (\textbf{A}, \%) and output tokens (\textbf{OT}, in thousands) of different models across four benchmarks. Results show that Reasoning Gates induce higher token consumption and yield the largest accuracy gap between reasoning and non-reasoning models.}
  \label{tab:gates-performance}

  \setlength{\tabcolsep}{3.1pt} 
  \begin{tabular}{@{}l cc cc cc cc @{}}
    \toprule
    \textbf{Model} 
    
      & \multicolumn{2}{c}{\parbox[c]{1.2cm}{\centering Murder \\ Mystery}} 
      & \multicolumn{2}{c}{\parbox[c]{1.3cm}{\centering Object \\ Placement}} 
      & \multicolumn{2}{c}{\parbox[c]{1cm}{\centering Flawed \\ Fiction}} 
      & \multicolumn{2}{c}{\parbox[c]{1.2cm}{\centering rRG \\ (Ours)}} \\
    
    \cmidrule(lr){2-3} \cmidrule(lr){4-5} \cmidrule(lr){6-7} \cmidrule(l){8-9}
     & \textbf{A(\%)} & \textbf{OT} & \textbf{A(\%)} & \textbf{OT} & \textbf{A(\%)} & \textbf{OT} & \textbf{A(\%)} & \textbf{OT} \\
    \midrule
    GPT-4o              & 70 & 0.3 & 48 & 0.01 & 58 & 0.2 & 37 & 0.4 \\
    Gemma3-27B         & 54 & 0.3 & 54 & 0.01 & 49 & 0.3 & 28 & 0.4 \\
    \cmidrule{1-9}
    \textbf{Average}    & 60 & 0.3 & 53 & 0.01 & 60 & 0.3 & \textbf{33} & \textbf{0.8} \\
    \midrule
    o3                  & 66 & 0.2 & 47 & 0.72 & 80 & 0.8 & 61 & 4.5 \\
    o3-mini             & 72 & 0.3 & 45 & 0.79 & 56 & 0.4 & 57 & 4.8 \\
    DeepSeekR1         & 73 & 0.3 & 53 & 1.5 & 74 & 1.1 & 37 & 5.0 \\
    Claude3.7S   & 72 & 1.2 & 50 & 1.0 & 67 & 0.3 & 50 & 3.4 \\
    Gemini2.5F    & 55 & 0.3 & 56 & 0.01 & 74 & 0.3 & 35 & 1.6 \\
    \cmidrule{1-9}
    \textbf{Average}    & 71 & 0.5 & 49 & 0.7 & 70 & 0.6 & \textbf{51} & \textbf{4.4} \\
    \bottomrule
  \end{tabular}
\end{table}

\paragraphbe{Humans against Web Agents.}
To test whether gig work can bypass rRGs, Table~\ref{tab:human_eval} compares human and agent performance on Reasoning Gate challenges. We recruited 10 participants who attempted the tasks without any LM assistance. Agents made attempts using two methods: the web interface and MCP. They succeeded in 10/10 web-interface attempts (while taking 3 mins and 46 secs on average) and 10/10 MCP attempts (while taking 24 secs on an average). Human participants succeeded in 2/10 attempts without internet access and 6/10 with internet (mean 11:30 and 11:00, respectively). Under a fixed time budget, these results suggest gig workers are unlikely to bypass the gate, showing resistance to cheap shortcuts (property \textbf{R1}).
\begin{figure}[ht]
\centering
\includegraphics[width=0.9\linewidth]{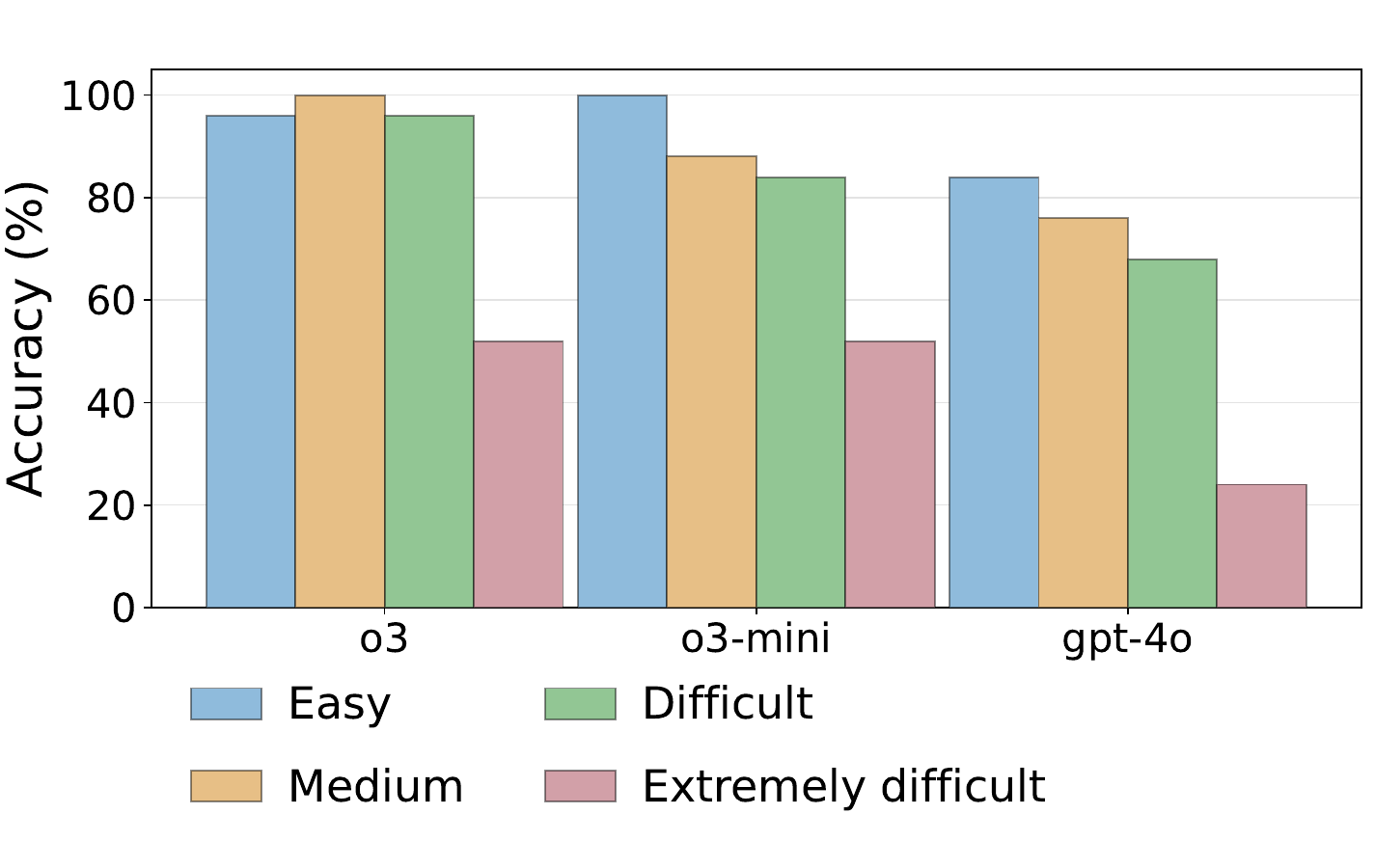}
\caption{\textbf{Performance by Difficulty.} Models show decreasing accuracy as task difficulty increases.}%
\label{fig:difficulty_based_performance}
\end{figure}

\paragraphbe{Offline Sample Generation Quality.}
\label{offiline_difficulty}
In Table~\ref{tab:icl_difficulty}, we show the quality of challenges generated to serve as ICL examples for large-scale offline generation.  We use o3-mini as our generator model $G_{icl}$ and use GPT-4o, o3-mini and o3 as our trusted models $\mathcal{A}^t$. Our results show that at any given difficulty level, the model's performance improves with an increase in reasoning capabilities. Furthermore, increasing the difficulty level leads to a consistent drop in performance across all models, highlighting the quality of difficulty control in generated challenges. Only o3 is able to solve some of the problems in the most difficult set, dropping the accuracy of GPT-4o and o3-mini to 0\%. This shows that the samples generated by $G_{icl}$ serve as good ICL examples for large-scale generation.

In Figure~\ref{fig:difficulty_based_performance}, we report the accuracy (\%) of different models on challenges of varying difficulty generated during the large-scale generation phase. Similar to ICL sample generation, o3 outperforms the other models across all difficulty levels, with one exception: in the easy challenge set, o3-mini solves one more challenge than o3. Overall, these results demonstrate the effectiveness of the difficulty control mechanism (property \textbf{C1}).

\paragraphbe{Comparison with Challenges in Existing Benchmarks.}
In Table~\ref{tab:gates-performance}, we compare Reasoning Gates (mixed difficulty challenges) with existing reasoning benchmarks on two properties: token consumption and the accuracy gap between reasoning and non-reasoning models. Reasoning Gate induces higher token usage than all other benchmarks for both reasoning models (up to 300 tokens) and non-reasoning models (up to 4.8k tokens), indicating that our challenges enable more effective throttling. We also observe that Reasoning Gates yield the largest gap in average accuracy between reasoning and non-reasoning models, whereas Murder Mystery and Object Placement datasets make reasoning models perform worse than non-reasoning models.

We hypothesize that the higher token consumption arises because, unlike reasoning challenges in other benchmarks, Reasoning Gates require the model to draw on multiple domains, whereas those challenges typically lie within a single domain. Because each clue can have many possible solutions, the language model must rely on context to identify the correct answer. As discussed earlier, we also hypothesize that the larger accuracy gap occurs because the final solution in Reasoning Gate must be an English word, making it easier for reasoning models to detect mistakes and backtrack.

\subsection{Adaptive Adversaries.}
In Table~\ref{tab:adaptive-adversaries}, we study the behavior of an adaptive adversary that employs many-shot ICL and fine-tuning, two popular task-specific learning strategies, to improve performance on the gates and decrease cost. Across all models, many-shot ICL improves performance by 3–5\%. Fine-tuning, by contrast, shows greater variance in success rates: for Llama and Gemma it performs worse than many-shot ICL, and for GPT-4o it even underperforms the baseline. We do not see any noticeable trend in number of tokens when comparing baseline output to fine-tuned models. We speculate that this occurs because fine-tuning may cause the adversary to overfit to characteristics of cached problems, yielding little improvement over the baseline. For larger models such as GPT-4o, this overfitting can potentially degrade generalization ability.


\begin{table}[ht]
\centering
\small
\caption{\textbf{Accuracy (\%) of evasion methods for reasoning gates.} Many-shot ICL yields modest improvements across all models (3-5\%), while fine-tuning produces mixed outcomes}
\label{tab:adaptive-adversaries}
\begin{tabular}{l@{\hspace{0.8cm}}ccc}
\toprule
\textbf{Models} &
\textbf{Baseline} &
\textbf{M-Shot ICL} &
\textbf{FT} \\
\midrule
GPT-4o   & 37\% & 40\%  & 34\%  \\
GPT-4.1 nano   & 21\% & 24\% & 26\%  \\
Llama-3.1-7B   & 7\% & 11\% & 10\%  \\
Gemma 3-4b   & 4\% & 9\% & 5\%  \\
\bottomrule
\end{tabular}
\end{table}  
\section{Discussion and Limitations}

\pbe{Availability of Resources for Deployment.}
Although Reasoning Gates provide strong asymmetry, the operational cost is nontrivial for small-scale resource providers, concentrating viable deployment among larger organizations. As with reCAPTCHA~\cite{plesner2024breaking} and Turnstile~\cite{cloudflareEmbedWidget}, we anticipate that firms with sufficient resources (e.g., LM Providers) will generate and maintain challenge banks and expose them as a service.

\pbe{Energy Consumption and Environmental Impact.}
The resource and energy demands of LMs have already led to serious environmental concerns~\cite{faiz2023llmcarbon}. We acknowledge that our framework increases LM consumption on both the client and provider sides. However, by analogy to proof-of-useful-work~\cite{ball2017proofs}, one could design useful Reasoning Gates in which generation and response compute is repurposed for real-world computational tasks.

\pbe{Limitations.} While major firms may have the resources to deploy the framework, current generation costs are higher than desired, reducing the incentive for large-scale deployment. We present preliminary results against adaptive attackers, but the impact of high-resource adversaries remains open. Finally, at scale we must study how difficulty settings affect latency and user experience across diverse users. We will address these challenges in future work.

\section{Conclusion}
The upcoming wave of LM-powered web agents will overload critical online
infrastructure, and traditional CAPTCHA-style defenses are no longer
sufficient. We formalized the web agent throttling problem, analyzed why
existing challenge problems fall short, and introduced the \emph{Reasoning
Gates} framework—automatically generated rebus puzzles that create
compute‐cost asymmetric problems to prevent DoS attacks while assessing an
agent’s capabilities. Our experiments demonstrate these gates can throttle
reasoning LMs including real web agents. We conclude by outlining the practical
opportunities and remaining challenges in deploying LM-generated throttling
mechanisms at scale.

\section*{Acknowledgments}

The work was partially supported by Schmidt Sciences SAFE-AI program and by NSF 2333965. 

\appendix
\vspace{-10pt}
\label{appendix:open_science}
\section{Ethical Considerations}


Our research studies “Reasoning Gates” for throttling LM web agents—lightweight, model-agnostic challenges that add user-side effort. The aim is to help operators reduce abusive automation while preserving access for legitimate users and tools. Stakeholders include end users (who may experience friction), service operators and content owners, agent developers/providers, and the broader public. Indirect stakeholders include accessibility and assistive-tech users, for whom we support verified-human bypasses, as well as researchers and policymakers who rely on transparent reporting. Evaluations ran on a custom website and MCP endpoints with real agents plus a small volunteer study; we did not probe or disrupt third-party services, and we logged only aggregate timing/accuracy without PII.

We recognize dual-use risk: gates could be misused to exclude benign automation or to unfairly block access of legacy users. To reduce harm, we use adjustable difficulty and pass criteria and cascade it with authentication and attestation to provide human-bypass options. For reproducibility without enabling misuse, we release specifications and evaluation code, alongside a precomputed challenge bank. We believe the benefits—clarifying trade-offs and guiding safer deployments—outweigh the potential harm that could be caused by the contributions in this paper.

\section{Open Science}
In alignment with the USENIX Security Open Science Policy, we make our artifacts publicly available. We share datasets to reproduce the presented results, and scripts to generate those datasets. The code can be found here \url{https://anonymous.4open.science/r/proof_of_reasoning_cleaned-4934/offline_generation.py}


\bibliographystyle{plain}
\bibliography{main}

\appendix
\appendix
\section{Prompts}
The ICL generator system prompt, the prover prompt, and the difficulty tuning command for ICL generation step are presented in Figures~\ref{fig:verifier_prompt}, \ref{fig:prover_prompt}, and~\ref{fig:difficulty_prompt}, respectively.

\label{appendix:verifier_prompt}
\begin{figure*}[ht!]
    \centering
    \begin{tcolorbox}[
        enhanced,
        colframe=purple!50!black,
        colback=purple!5,
        coltitle=white,
        colbacktitle=purple!50!black,
        width=\textwidth,
        arc=3mm,
        boxrule=0.8mm,
        drop shadow,
        title=\small\bfseries ICL Generator System Prompt,
        fonttitle=\bfseries\small
    ]
    {\small

Can you create a so called "reasoning rebus". Can you create a puzzle (or gate) that tests knowledge across all the disciplines to hide text. Furthermore, make it layered, that only by solving previous part (called gates) you will get to the next word or gate to solve, so it should be very iterative. You will first print entire gate and then print the solution like -
s
1. Gate = \{The entire rebus\} \\
Solution = \{The solution to the above rebus\}

Only provide solution once you are done printing the rebus and don't provide any clues.

While asking for the new gate the user will provide previous few gates and the difficulty request they gave the LM while generating those gates. ......

Previous Rebuses - \\
Gate - \{The generated gate\} \\
Solution - \{\} \\
difficulty request - \{\} \\
... \\
Gate - \{The generated gate\} \\
Solution - \{\} \\
difficulty request - \{\} \\
new difficulty request - \{\}

If no previous gate or difficulty levels are specified then that means it is the first puzzle. In that scenario, the user request will look like - \\
Previous Rebuses - None \\
Difficulty - Generate first gate \\
........ \\
and don't mention the user difficulty instruction in your output. 

Here are a few examples of generations -

USER INPUT - \\
Previous Rebuses - None \\
new difficulty request - Generate first gate

Previous Gate Number = 0 \\
Domains = [Planetary Geology, Paleobotany, Cake Decorating, Packaging Engineering, Naturopathy] \\
Words = [enosis, zogan, music, toled]

OUTPUT - \\
1. Gate = Begin your journey by deciphering these interdisciplinary clues: \\
• Planetary Geology: Identify the red planet, renowned for its dusty, iron-oxide landscape and named after the Roman god of war. Extract its first letter. \\
• Paleobotany: In the study of ancient plant life, uncover the name of an enigmatic seed fern mentioned in obscure Carboniferous texts as “Ulospora.” Take the initial letter of this genus. \\
...... \\
Solution = music

… (etc.) …

3. Gate = Unravel the clues below by determining the correct answer for each discipline and then extracting the specified letter to reveal the hidden word: \\
• Packaging Engineering: Identify the process that uses controlled heat to seal flexible packages. Record the first letter of its name. \\
......
Solution = technical
    }
    \end{tcolorbox}
    \caption{\small Full specification for generating layered reasoning rebus gates.}
    \label{fig:verifier_prompt}
\end{figure*}

\label{appendix:prover_prompt}
\begin{figure*}[ht!]
    \centering
    \begin{tcolorbox}[
        enhanced,
        colframe=purple!50!black,
        colback=purple!5,
        coltitle=white,
        colbacktitle=purple!50!black,
        width=\textwidth,
        arc=3mm,
        boxrule=0.8mm,
        drop shadow,
        title=\small\bfseries Prover Prompt,
        fonttitle=\bfseries\small
    ]
    {\small

You are a puzzle solving agent who will be given puzzles "gate" and you will solve the puzzle to the best of your abilities. While you can generate reasoning before final solution, put the solution for the provided challenge as python dictionary in the end structured like this {“Gate n”: solution string}. Where the key in the dictionary is the assigned number of the gate you are answering in that round of the interaction.

Here is an example - 

PROVIDED GATE - 

gate =

For each clue below, determine the answer and then record its first letter. In sequence, these letters will reveal a hidden word:

• Computer Science – Subfield 39: Identify the popular high-level programming language known for its snake logo. (Take the first letter of your answer.)

• Medicine – Subfield 43: Name the condition characterized by an abnormal accumulation of fluid in body tissues. (Record its starting letter.)

• Physics – Subfield 8: Identify the universal constant denoted by the letter "c" that represents the speed of light. (Take its letter as your answer.)

• Medicine – Subfield 15: What is the name of the joint connecting the thigh bone to the lower leg? (Extract the first letter.)
• Physics – Subfield 42: Determine the term for the rate of change of velocity. (Provide its initial letter.)

• Computer Science – Subfield 27: Identify the data structure that represents relationships using nodes (vertices) and connections (edges). (Record the first letter.)

• Mathematics – Subfield 17: What is the term for a statement that asserts the equality of two expressions? (Take its starting letter.)

Solution = 

OUTPUT - 

To solve the puzzle for Gate 7, let's find the first letter of each answer based on the given clues:

1. **Computer Science – Subfield 39:** The popular high-level programming language known for its snake logo is **Python**. First letter: **P**

2. **Medicine – Subfield 43:** The condition characterized by an abnormal accumulation of fluid in body tissues is known as **Edema**. First letter: **E**

3. **Physics – Subfield 8:** The universal constant denoted by the letter "c" that represents the speed of light is **c** itself (commonly known as the speed of light in equations). Therefore, the letter is **C**

4. **Medicine – Subfield 15:** The joint connecting the thigh bone to the lower leg is the **Knee**. First letter: **K**

5. **Physics – Subfield 42:** The term for the rate of change of velocity is **Acceleration**. First letter: **A**

6. **Computer Science – Subfield 27:** The data structure that represents relationships using nodes (vertices) and connections (edges) is a **Graph**. First letter: **G**

7. **Mathematics – Subfield 17:** The term for a statement that asserts the equality of two expressions is an **Equation**. First letter: **E**
Combining these initial letters, we get the hidden word: **PECKAGE**

The solution: {"Gate": "PECKAGE"}

    }
    \end{tcolorbox}
    \caption{\small Full specification for generating layered reasoning rebus gates.}
    \label{fig:prover_prompt}
\end{figure*}

\label{appendix:difficulty_prompt}
\begin{figure*}[ht!]
    \centering
    \begin{tcolorbox}[
        enhanced,
        colframe=purple!50!black,
        colback=purple!5,
        coltitle=white,
        colbacktitle=purple!50!black,
        width=\textwidth,
        arc=3mm,
        boxrule=0.8mm,
        drop shadow,
        title=\small\bfseries Difficulty Tuning Command for ICL Generation Step,
        fonttitle=\bfseries\small
    ]
    {\small
"First":"Generate first gate", 

"Increase":"Generate another gate and increase difficulty",

                         "Maintain":"Good, maintain difficulty and ambiguity generate another gate", 
                         
                         "Reduce":"generate new problems with reduced difficulty"
    }
    \end{tcolorbox}
    \caption{\small Full specification for generating layered reasoning rebus gates.}
    \label{fig:difficulty_prompt}
\end{figure*}

\end{document}